# AED-Net: An Abnormal Event Detection Network

Tian Wang [a], Zichen Miao [a], Yuxin Chen[a], Yi Zhou[b], Guangcun Shan*[c], Hichem Snoussi[d]

a School of Automation Science and Electrical Engineering, Beihang University, Beijing 100191, China

b Department of Electronic Engineering, Dalian Maritime University, Dalian 116026, China

c School of Instrumentation Science and Opto-electronic Engineering & International Research Institute for Multidisciplinary Science, Beihang University, Beijing 100191, China

d Institute Charles Delaunay-LM2S-UMR STMR 6281 CNRS, University of Technology of Troyes, Troyes 10010, France

*Corresponding author.
 E-mail address: gcshan@buaa.edu.cn

**ABSTRACT** It is challenging to detect the anomaly in crowded scenes for quite a long time. In this paper, a self-supervised framework, abnormal event detection network (AED-Net), which is composed of PCAnet and kernel principal component analysis (kPCA), is proposed to address this problem. Using surveillance video sequences of different scenes as raw data, PCAnet is trained to extract high-level semantics of crowd's situation. Next, kPCA, a one-class classifier, is trained to determine anomaly of the scene. In contrast to some prevailing deep learning methods, the framework is completely self-supervised because it utilizes only video sequences in a normal situation. Experiments of global and local abnormal event detection are carried out on UMN and UCSD datasets, and competitive results with higher EER and AUC compared to other state-of-the-art methods are observed. Furthermore, by adding local response normalization (LRN) layer, we propose an improvement to original AED-Net. And it is proved to perform better by promoting the framework's generalization capacity according to the experiments.

## 1 Introduction

As a step after static image, studies on video has attracted more and more researchers' attention in computer vision community in recent years. Lately, researches like object tracking[1-3], gait recognition[4, 5] and activity recognition[6-8] achieved competitive results and all have promising future.

Abnormal event detection, a task to detect the specific frames containing an anomaly, is also among the hottest research issues in the video field. Comparing to the tasks discussed previously, abnormal event detection has more significance on national security and people's livelihoods. With the modernization of the society, an increasing number of surveillance cameras are deployed in different places, producing a large quantity of video every second, which are impossible for a human to handle with and figure out the abnormal events in them. However, even once miss of anomaly could cause unbearable loss. Thus, it is important to construct an automatic video abnormal detector dealing with thousands of hundreds of videos frames and alerting people for a timely and effective response when an anomaly happens.

There are many difficulties in anomaly detection described in [9]. Though we can easily list a couple of kinds of abnormalities in a specific scene, such as a cart or a biker in the crowd scene, it is impractical to enumerate all possible abnormal events in one scene, meaning there are countless positive classes in this classification task. Furthermore, due to lack of abnormal samples, i.e., video frames including abnormal events, the training set is severely imbalanced, implying that it is infeasible to train a model for multi-classes classification. All the difficulties above indicate that the anomaly detection task is a hardly handled one-class classification problem.

There are some methods that have been suggested to deal with abnormal event detection. A method was proposed in[10] based on histograms of the optical flow orientation descriptor. As the handcrafted feature descriptor was constructed based on the experience of the human, it did not represent the feature in a training process. Thus, it performs worse than present deep learning methods. Lately, as described in[11, 12], deep learning methods have been developed largely due to the availability of big data and efficient hardware. They are applied intensively in computer vision field and has achieved great results. Researchers in [13] uses convolution neural network (CNN) for defect detection in product quality control. However, the original CNN for face recognition is not applicable to this task, because its training needs samples of different classes. Considering the success of PCAnet [14] in image classification, [39] proposed a PCAnet based method to extract information from raw image for anomaly detection with a one-class classifier constructed on clustering algorithm. However, it has natural limitations due to K-medoids clustering algorithm which is difficult to deal with the high-dimensional feature extracted by PCAnet. In this paper, we propose a self-supervised network, abnormal event detection network (AED-Net), to deal with this video anomaly detection task, as only normal samples are provided as training data. Since PCAnet has been proved its ability to extract feature as an unsupervised model, it suits out self-supervised AED-Net. Furthermore, a one-class classifier which handles the extracted high-dimension feature is used to determine the abnormality of frames.

To be specific, this new self-supervised network uses the optical flow maps as the input, because these maps represent motion in a better way. Then we can extract high-level semantics of crowd's situation from PCAnet. Next, a simple but effective one-classifier, kernel principal component analysis (kPCA) [35] is used to classify the high-dimensional feature. Having merits from both networks, AED-Net is trained to understand each frame and conduct the detection. More importantly, Local Response Normalization layer (LRN layer), a trick used in CNN to aid generalization, is combined with the AED-Net for improvement. It should be worth noting that this new network can be trained with unlabeled data and perform better by comparing with the state-of-the-art methods in an abnormal detection task and our new self-supervised network can effectively detect abnormal events even in crowded situations, which improves the detection results according to the experiments tested on the public UMN dataset and UCSD dataset.

The rest of this paper is organized as follows. In Section2, related works are reviewed briefly. Section 3 reviews the basic algorithms of our framework, PCAnet and kPCA. Then the whole architecture of AED-Net will be elaborated in Section 4, and our improvement to it is also introduced in this section. Then, in Section 5, the experimental results of datasets, UMN [30] and UCSD [31] are illustrated and discussed. Finally, conclusions are presented in Section 6. The introduction of notations used in this paper is shown in table 1.

## 2 Related work

In general, traditional methods for anomaly detection can be divided into two major classes. The first class is based on trajectory. This method has been used broadly in abnormal events detection[15-18]. In [19-21], authors extract the trajectory of normal events to indicate normal modes, and trajectories that differ from the normal patterns are considered abnormal. However, the occlusion between moving objects affects the effectiveness of this method when it's applied to crowded occasions. To tackle this problem, a new model is suggested in [22] to deal with the interrelatedness of people's behavior and to ameliorate the representation of objects interaction. And in [23], a discrete transformation is utilized to develop a reliable multi-target tracking algorithm that associates the objects in different frames. However, the occlusion problem affects the results so much that the methods above do not address this issue in an effective way. Hence, the tracking strategy is not adopted in our work.

The method based on spatiotemporal is another category. Some promising researches have been proposed. In [24], researchers propose a feature descriptor, covariance matrix, which encodes optical flow and partial derivatives of adjacent frames. In [25-28], authors model motion patterns with histograms of pixel changes. In[29-31], distributions of optical flow are used as the basic features, and then models for detecting abnormal events are built based on optic flow features. [32] proposed an approach to estimate the interaction between moving objects. A study [9] use a detector combining time and space anomalies. The wavelet transform used in image processing can also be utilized to analyze the motion [33, 34]. The delicate feature descriptors were designed manually, tending to only work well under specified conditions. In our work, the feature is extracted by a self-supervised network.

With the rapid development of deep learning, it has achieved outstanding results in the field of abnormal event detection recently. Unlike the features of manual design, features extracted by deep learning network are obtained through a learning process. One self-supervised learning method is proposed, only normal samples are learning in the proposed AED-Net for abnormal event detection.

## 3 Self-supervised feature extraction and anomaly detection

Self-supervised learning is a learning paradigm that there is no external supervised information, i.e., the labels, as the ground truth beyond the data itself. Under this paradigm, the self-supervised learning method just adopts the raw data as the material for training, which means that the model learns to extract latent supervised information in the data. The categories of the data are not employed in the training process.

The self-learning model is applicable to the anomaly detection task. Since we can only use the normal data to train the model, there is no external supervised information given to the model. Thus, the model must fully understand what is normal datum from the input video clips and use it as supervised information to tune its parameters.

**Table 1** A description of notations used in this paper

| Variable | Description |
| --- | --- |
| $S$ | Raw surveillance video frame |
| $I$ | Input of PCAnet (optical flow map) |
| $k_1, k_2$ | Size of patches in PCAnet |
| $X$ | Matrix consists of all patches from an optical flow map at stage one of PCAnet |
| $S_i$ | Matrix containing all matrix $\bar{X}$ at $i$-th stage of PCAnet |
| $K_j^i$ | $j$-th filter of $i$-th stage of PCAnet |
| $C$ | Outputs of the first stage of PCAnet |
| $Y$ | Matrix consists of all patches from an optical flow map at stage one of PCAnet |
| $O$ | Outputs of the second stage of PCAnet |
| $T$ | Integer-valued image after binarizing and encoding outputs of stage 2 |
| $\mathcal{F}$ | The final feature of PCAnet |
| $F$ | Inputs feature of kPCA classifier |
| $\mathcal{M}(F_i)$ | Feature F mapped into higher-dimensional space |
| $\kappa(F_i, F_j)$ | Scalar product of $\mathcal{M}(F_i)$ and $\mathcal{M}(F_j)$ |
| $W_j$ | Eigenvectors of covariance matrix of mapped feature in higher-dimensional space |
| $V$ | Kernel matrix, $V_{i,j} = \kappa(F_i, F_j)$ |
| $\bar{V}$ | Kernel matrix, $\bar{V}_{i,j}$ is scalar product of $\bar{\mathcal{M}}(I_i)$ and $\bar{\mathcal{M}}(I_j)$ |
| $R$ | Reconstruction error, i.e., abnormality score |
| $\alpha$ | eigenvectors of kernel matrix $\bar{V}$ |
| $p_i$ | output value on the $i$-th feature map |
| $q_i$ | normalized output of $p_i$ |
| $\delta, n, \beta, \gamma$ | Hyperparameters of local response normalization |

### 3.1. PCAnet for feature extraction

Both traditional and deep learning methods has been applied for features extraction from video frames. In [10], the global optical flow descriptor is used as the feature. However, optical flow only contains the low-level motion information in the frames, high-level information feature like people's running pattern, or how many people are in the frame, cannot be represented by it. Thus, deep learning method is used to deal with this high-level feature extraction problem. The most popular model is CNN, which stack layers and extract deeper and deeper feature step by step. However, this particular model needs strong external supervised information, which is not provided in our task. Thus, we choose the PCAnet [34], an equivalent model in feature extraction without the need of external supervised information and utilizing the power of deep learning.

PCAnet[14] is a deep learning network proposed under the prevailing trend of deep learning. Though it is simple compared to other popular deep learning networks, such as deep convolution neural network (CNN), it is capable enough to handle challenging tasks like face recognition. Thus, the model is used considering its efficiency and competitive ability in feature extraction.

PCAnet is a cascaded linear network. A typical two-stage PCAnet architecture is shown in Fig.1. Because it is inspired by CNN, each stage of PCAnet is consists of independent PCA filter bank that is needed to be learned and perform feature extraction work. Feature maps in the first stage are linearly cascaded to next stage for extracting higher-level features. The performance corresponding to the number of stages discussed by Chan et al. [14] shows that though two-stage network performs better than the one-stage network, networks with more than two stages don't have many advantages over two-stage network, so two-stage PCAnet is enough for the task at hand for the benefit of computation efficiency.

As talked above, the two-stage PCAnet is used to extract features. In training phase, at the beginning of stage one, an optical flow map $I_i$ with a shape of $h \times w$ will be sampled around each pixel to small patches of a size of $k_1 \times k_2$, which is shown in Fig.2 with gray arrows. Then the samples, $patch(x_1), patch(x_2), \ldots, patch(x_{(h-k_1+1)\times(w-k_2+1)})$, will be vectorized and compose sample matrix $X = [x_1, x_2, \ldots, x_{(h-k_1+1)\times(w-k_2+1)}]$. Next, we perform mean subtraction to $X$ and get $\bar{X}$. All the descriptions of notations used in this paper are shown in Table 1.

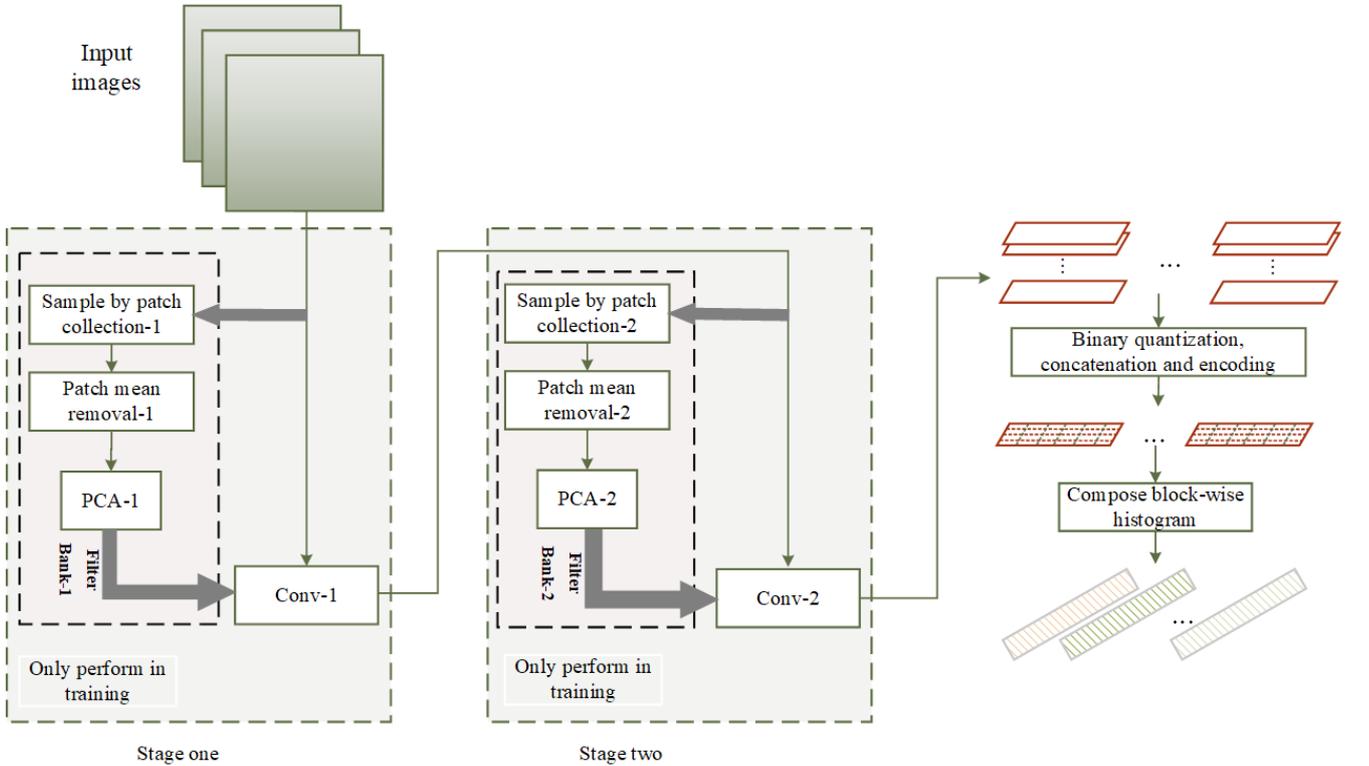

**Fig. 1.** The typical structure of two-stage PCAnet used in our method

For $N$ input optical flow maps, $I = \{I_1, I_2, \ldots, I_N\}$, PCAnet initially samples them and get:
$$S_1 = [\bar{X}_1, \bar{X}_2, \ldots, \bar{X}_N] \in \mathbb{R}^{k_1 k_2 \times N(h-k_1+1)\times(w-k_2+1)} \tag{1}$$
Then, PCAnet computes $L_1$ convolution kernels based on $I$ by implementing principal component analysis, as shown by gray arrows in Fig.2, and get:
$$K_l^1 = vec2mat_{k_1,k_2}\left(s_l(S_1 S_1^T)\right) \in \mathbb{R}^{k_1 \times k_2}, \quad l = 1,2,\ldots,L_1, \tag{2}$$
where $s_l(S_1 S_1^T)$ denotes $l$-th principal eigenvector of $S_1 S_1^T$, and vec2mat(·) map a vector from $\mathbb{R}^{k_1 k_2}$ to a matrix $M \in \mathbb{R}^{k_1 \times k_2}$. At the end of stage one, the convolution operation is performed to extract features:
$$C_i^l = I_i * K_l^1, \quad i = 1,2,\ldots,N, \tag{3}$$
where * denotes 2-D convolution, $C_i^l$ means the $l$-th feature map of the $i$-th input $I_i$, and the number of outputs in stage one is $L_1N$. Note that the boundary of $I_i$ is zero-padded to entail the outputs have the same size of the input, i.e., $h \times w$. As it is implied, in the test phase, PCAnet will directly perform convolution operation on inputs $I$ using kernels obtained from training phase.

The second stage was conducted in almost the same way as the first one. In training phase, each input $C_i^l$ of $C$ will be sampled to patches. And these patches will be vectorized and compose matrix $S_2$ after performing mean subtraction,

$$S_2 = [\bar{Y}_1^1, \ldots, \bar{Y}_1^{L_1}, \bar{Y}_2^1, \ldots, \bar{Y}_2^{L_1}, \ldots, \bar{Y}_N^1, \ldots, \bar{Y}_N^{L_1}] \in \mathbb{R}^{k_1 k_2 \times L_1 N(h-k_1+1) \times (w-k_2+1)}, \quad (4)$$

where $\bar{Y}_j^l = [\bar{y}_{i,l,1}, \bar{y}_{i,l,2}, \ldots, \bar{y}_{i,l,(h-k_1+1) \times (w-k_2+1)}] \in \mathbb{R}^{k_1 k_2 \times (h-k_1+1) \times (w-k_2+1)}$, means the sample matrix of $C_i^l$. Then we compute convolution kernels in stage two,

$$K_m^2 = vec2mat_{k_1,k_2}\left(s_m(S_2 S_2^T)\right) \in \mathbb{R}^{k_1 \times k_2}, \quad m = 1,2,\ldots,L_2 \quad (5)$$

Finally, we get outputs of stage two by convolution,

$$O_i^{l,m} = C_i^l * K_m^2, \quad l = 1,2,\ldots,L_1, \quad m = 1,2,\ldots,L_2, \quad i = 1,2,\ldots,N \quad (6)$$

The number of outputs in stage two is $L_2 L_1 N$.

After stage two, we would binarize the output by Heaviside step function $H(\cdot)$, assigning one for positive entries and zero for zero or negative entries. This enables the network to have nonlinearity. Thus, the network is capable to capture high-level semantics in the optical flow maps. For each of these $L_1$ inputs of the second stage $C_i^l$, it has $L_2$ real-valued outputs in the second stage $O_i^{l,m}$, $(m = 1,2,\ldots,L_2)$. Around each pixel, there are $L_2$ binary bits and we can view them as a decimal number, converting the $L_2$ outputs $O_i^{l,m}$ to a single integer-valued image:

$$T_i^l = \sum_{m=1}^{L_2} 2^{m-1} H(O_i^{l,m}) \quad (7)$$

Finally, the output features of PCAnet are block histograms (with $2^{L_2}$ bins) computed based on all $T_i^l$. Note that, one histogram does not represent the whole $T_i^l$, but a region of it. To do so, $T_i^l$ is partitioned into $B$ blocks and then used to calculate histogram. histogram is computed in each block. And then all histograms are concatenated into one vector, Bhist($T_i^l$). For single input optical image $I_i$, the feature of it is:

$$\mathcal{F}_i = [Bhist(T_i^1), \ldots, Bhist(T_i^{L_1})]^T \in \mathbb{R}^{(2^{L_2})L_1 B} \quad (8)$$

The local block can be either overlapping and non-overlapping. The latter setting is beneficial for detection except for face detection [14], so in this paper, it is set to non-overlapping. Besides the overlapping choice, the hyper-parameters of the PCAnet also include the filter size $k_1, k_2$, the number of filters in each stage $L_1 L_2$, and the block size for local histograms.

### 3.2. A Self-supervised learning method for anomaly detection: kPCA

For the reason that we can only utilize video sequence of normal scenes, and we need to distinguish normal frames from abnormal ones which are previously unknown, it's appropriate to class this task as one-class classification.

The common idea in one-class classification task is to train a classifier that encloses the training data, i.e., the normal data, and in such way separate abnormal data from normal data. The support vector domain description (SVDD) classifier is a good example of this method. However, this classifier often generates a too large decision boundary to perform well enough. Using Gaussian process priors, [35] build model for one-class classification. It uses different measures derived from Gaussian process regression and approximate Gaussian classification. However, it strongly relies on hyperparameter-tuning of the re-parameterized kernel function.

On the contrary, learning the distribution of data which is usually non-linear, kPCA classifier[36] can generate a decision boundary smoothly follow the distribution of data and tends to classify more accurately.

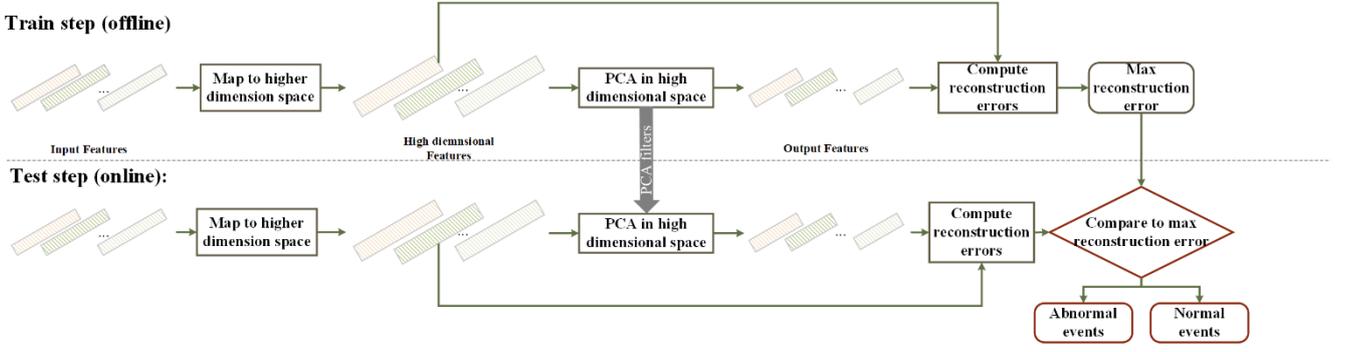

**Fig. 2.** The structure of one-class classifier: kPCA

The structure of kPCA classifier is shown in Fig.2. The essential idea of this one-class classifier is that features of normal frames have the similar distribution, while the distribution of features of abnormal frames is much different. Thus, after using PCA filters computed based on training features, i.e., normal features, to do PCA on both normal features and abnormal features, we could observe a clear difference of reconstruction error between normal features and abnormal features. Then the classification can be conducted according to this disparity.

As discussed by Hoffmann [36], PCA cannot capture the non-linear structure of input. Hence, kPCA is introduced to overcome this drawback, which maps input $F_i \in \mathbb{R}^m$ to feature in higher-dimensional space: $\mathcal{M}(F_i) \in \mathbb{R}^n$ (n>m). PCA is then performed in the feature space. Computation here only requires scalar product of $\mathcal{M}(F_i)$, i.e., $(\mathcal{M}(F_a) \cdot \mathcal{M}(F_b))$. And the scalar product is further be replaced by kernel function $\kappa(F_i, F_j)$ to perform the same task. Here, the kernel function uses Gaussian kernel $\kappa(F_i, F_j) = \exp(-\frac{\|F_i - F_j\|^2}{2\sigma^2})$. Furthermore, we get $\overline{\mathcal{M}}(F_i)$ from $\mathcal{M}(F_i)$ by performing mean subtraction, which could further represent $W_j$, eigenvectors of covariance matrix in higher-dimensional space. Thus, $W_j$ can be expressed by $\mathcal{M}(F_i)$ as,

$$W_j = \sum_{i=1}^{N} \alpha_i^j \left( \mathcal{M}(F_i) - \frac{1}{N} \sum_{k=1}^{N} \mathcal{M}(F_k) \right), \quad (9)$$

It turns out that $\alpha^j$, $(\alpha^j = [\alpha_1^j, \alpha_2^j, \ldots, \alpha_N^j]$, $j = 1,2,\ldots,q)$ is an eigenvalue of kernel matrix $\overline{V}$. Each component of $\overline{V}$, i.e. $\overline{V}_{ij}$, is scalar product of $\overline{\mathcal{M}}(F_i)$ and $\overline{\mathcal{M}}(F_j)$. Similarly, each component of kernel matrix $V$, i.e., $V_{ij}$ is scalar product of $\mathcal{M}(F_i)$ and $\mathcal{M}(F_j)$. Thus,

$$\overline{V}_{ij} = V_{ij} - \frac{1}{N} \sum_{a=1}^{N} V_{ia} - \frac{1}{N} \sum_{a=1}^{N} V_{aj} + \frac{1}{N^2} \sum_{a,b=1}^{N} V_{ab}, \quad (10)$$

According to Hoffmann [36], given feature $F_z$, the reconstruction error is calculated in feature space,

$$R(I_z) = \left( \overline{\mathcal{M}}(F_z) \cdot \overline{\mathcal{M}}(F_z) \right) - \left( \left( W\overline{\mathcal{M}}(F_z) \right) \cdot \left( W\overline{\mathcal{M}}(F_z) \right) \right), \quad (11)$$

where $W = (W_1; W_2; \ldots; W_q)$. Then, the equation above can be expressed more clearly as,

$$R(F_z) = \|\overline{\mathcal{M}}(F_z)\|^2 - \sum_{j=1}^{q} \left( \overline{\mathcal{M}}(F_z) \cdot W_j \right)^2$$

$$= V_{zz} - \frac{2}{N} \sum_{a=1}^{N} V_{za} + \frac{1}{N^2} \sum_{a,b=1}^{N} V_{ab} - \sum_{j=1}^{q} \left( P_j(F_z) \right)^2$$

$$= 1 - \frac{2}{N} \sum_{a=1}^{N} V_{za} + \frac{1}{N^2} \sum_{a,b=1}^{N} V_{ab} - \sum_{j=1}^{q} \left( P_j(F_z) \right)^2, \quad (12)$$

In equation above, $P_j(F_z)$ has expression as below,

$$P_j(F_z) = \bar{\mathcal{M}}(F_z) \cdot W_j$$

$$= [\mathcal{M}(F_z) - \frac{1}{N}\sum_{a=1}^{N}\mathcal{M}(F_a)] \cdot [\sum_{i=1}^{N}\alpha_i^j \mathcal{M}(F_i) - \frac{1}{N}\sum_{i,b=1}^{N}\alpha_i^j \mathcal{M}(F_b)]$$

$$= \sum_{i=1}^{N}\alpha_i^j \left[ V_{zi} - \frac{1}{N}\sum_{a=1}^{N}V_{ia} - \frac{1}{N}\sum_{a=1}^{N}V_{zb} + \frac{1}{N^2}\sum_{a,b=1}^{N}V_{ab} \right], \tag{13}$$

Hence, we obtain the desired form of measurement $R(I_z)$ to detect the anomaly.

The hyper-parameters in this classifier are the number of eigenvectors $q$, and kernel width σ. Their values depend on the specific experiment environment.

Finally, given an input $X$ and extracted feature $F_x$, we define the classifier as,

$$\text{status}(X) = \begin{cases} anomaly & R(F_x) > threshold \\ normality & R(F_x) \leq threshold \end{cases} \tag{14}$$

The threshold above is the maximum reconstruction error computed in training phase, as shown in Fig.2.

## 4 Proposed abnormal event detection network

Given the task of anomaly detection in video frames, we propose AED-Net, an integral self-supervised detection framework based on self-supervised learning method which trains the normal data. To perform feature extraction task based on input video frames, PCAnet, an effective network is adopted. Then for one-class classification, we use kPCA, a particular one-class classifier to determine the abnormality of the frames.

### 4.1 Optical flow computation

Initially, we get raw video frames, $S$. To detect the abnormal events in those frames, moving area should be firstly separated in $S$ from static background for simplifying the detection task. Optical flow, which represents the motion field between frames[37], is applicable to this motion extraction requirement.

Horn-Schunck's method (H-S method)[38] can compute optical flow. Considering constraints of pixel values consistency as well as flow variety across image, this method constructs an energy function and optimizes it to get optical flow, $u$ and $v$[38]. $u$ and $v$ are horizontal and vertical component of the optical flow. And constraint of smoothness is added in the function to mitigate the aperture problem. The proposed energy function is,

$$E = \iint [(I_x u + I_y v + I_t)^2 + \alpha^2(\|\nabla u\|^2 + \|\nabla v\|^2)]dxdy \tag{15}$$

in which $E$ is global energy. $I_x$, $I_y$, and $I_t$ are pixel values across width direction, height direction and time direction. $\alpha$ is the hyperparameter controlling the smooth term.

Then, in order to process the optical flow feature like processing images, we visualize optical flow $u, v$ and get optical flow maps, $I$, using Munsell Color System.

### 4.2. AED-Net

Intuitively, the anomaly detection task in our proposed AED-Net is to assign a score indicating abnormality to each frame of video. And during a training phase, the largest reconstruction error should be set as threshold for the anomaly detection. Thus, in testing phase, the abnormality of test frames can be determined by comparing the score of test frames to the threshold. To fulfill this task, we incorporate PCAnet and kPCA together and build AED-Net.

The framework of our proposed AED-Net is shown in Fig.3. The proposed algorithm of the AED-Net is shown in Alg.1. Firstly, optical flow maps, $I$, are used as input of the whole framework for training and testing. Then, PCAnet model is trained to learn to extract high-level information that better represents the situation of the scenes from the spatial-temporal features. Finally, utilizing the block-wise histograms as classification features extracted by PCAnet, kPCA is trained to learn non-linear data distribution of normal scenes and determine the max normality score as the threshold computing by reconstruction error.

During test time, to minimize the influence of frames that carry little relevant information, the foreground detection is firstly performed and frames in test video clip that contain few people are removed. Then k block-wise features are extracted by the PCAnet trained previously, and a test score is computed for every frame by kPCA. Finally, the test score is used to determine whether the frame is abnormal by comparing to the max normality score.

---

**Alg.1 AED-Net**

**Input:** optical flow maps $I = \{I_1, I_2, ..., I_N\}$

**Output:** Threshold of max normality score $threshold$, $K^1$, $K^2$, $\alpha$

1. **for** $i = 1,2,...,N$ **do**
2.    Sample $I_i$ by patches and get $X_i$ by vectorizing and concatenating all patches
3. **end for**
4. $\overline{X_i} = X_i - \frac{1}{N}(\sum X_i)$
5. $S_1 = [\overline{X_1}, \overline{X_2}, ..., \overline{X_N}]$
6. $K^1 = \underset{K^1}{\operatorname{argmin}} \|S_1 - K^1(K^1)^T S_1\| \quad s.t. K^1(K^1)^T = I_{L_1}$
7. $C = I * K^1, \quad C = \{C_i^l\} \, i = 1,2,...,N. \, l = 1,2,...,L_1$
8. **for** $i = 1,2,...,N, j = 1,2,...,L_1$ **do**
9.    Sample $C_i^j$ by patches and get $Y_i^j$ by vectorizing and concatenating all patches
10. **end for**
11. $\overline{Y_i^j} = Y_i^j - mean(Y_i^j)$
12. $S_2 = [\overline{Y_1^1}, ..., \overline{Y_1^{L_1}}, \overline{Y_2^1}, ..., \overline{Y_2^{L_1}}, ..., \overline{Y_N^1}, ..., \overline{Y_N^{L_1}}]$
13. $K^2 = \underset{K^2}{\operatorname{argmin}} \|S_2 - K^2(K^2)^T S_2\| \quad s.t. K^2(K^2)^T = I_{L_2}$
14. $O_i^{l,m} = C_i^l * K_m^2, \quad l = 1,2,...,L_1, \quad m = 1,2,...,L_2, \, i = 1,2,...,N$
15. $T_i^l = \sum_{m=1}^{L_2} 2^{m-1} H(O_i^{l,m})$
16. $\mathcal{F}_i = [Bhist(T_i^1), ..., Bhist(T_i^{L_1})]^T$
17. **for** $i = 1,2,...,N,$ **do**
18.   **for** $j = 1,2,...,N,$ **do**
19.     $V_{i,j} = \kappa(\mathcal{F}_i, \mathcal{F}_j)$
20.   **end for**
21. **end for**
22. $\overline{V}_{ij} = V_{ij} - \frac{1}{N}\sum_{a=1}^{N} V_{ia} - \frac{1}{N}\sum_{a=1}^{N} V_{aj} + \frac{1}{N^2}\sum_{a,b=1}^{N} V_{ab}$
23. $\alpha = \underset{\alpha}{\operatorname{argmin}} \|\overline{V} - \alpha \alpha^T \overline{V}\| \quad s.t. \alpha \alpha^T = I_q$
24. Initialize threshold=0.
25. **for** $i = 1,2,...,N$ **do**
26.   $R(\mathcal{F}_i) = 1 - \frac{2}{N}\sum_{a=1}^{N} K_{za} + \frac{1}{N^2}\sum_{a,b=1}^{N} K_{ab} - \sum_{j=1}^{q}\left(P_j(\mathcal{F}_i)\right)^2$, and
27.   $P_j(\mathcal{F}_i) = \sum_{i=1}^{N} \alpha_i^j \left[K_{zi} - \frac{1}{N}\sum_{a=1}^{N} K_{ia} - \frac{1}{N}\sum_{a=1}^{N} K_{zb} + \frac{1}{N^2}\sum_{a,b=1}^{N} K_{ab}\right]$
28.   **if** $R(\mathcal{F}_i) > threshold,$ **do**
29.     $threshold = R(\mathcal{F}_i)$
30.   **end if**
31. **end for**

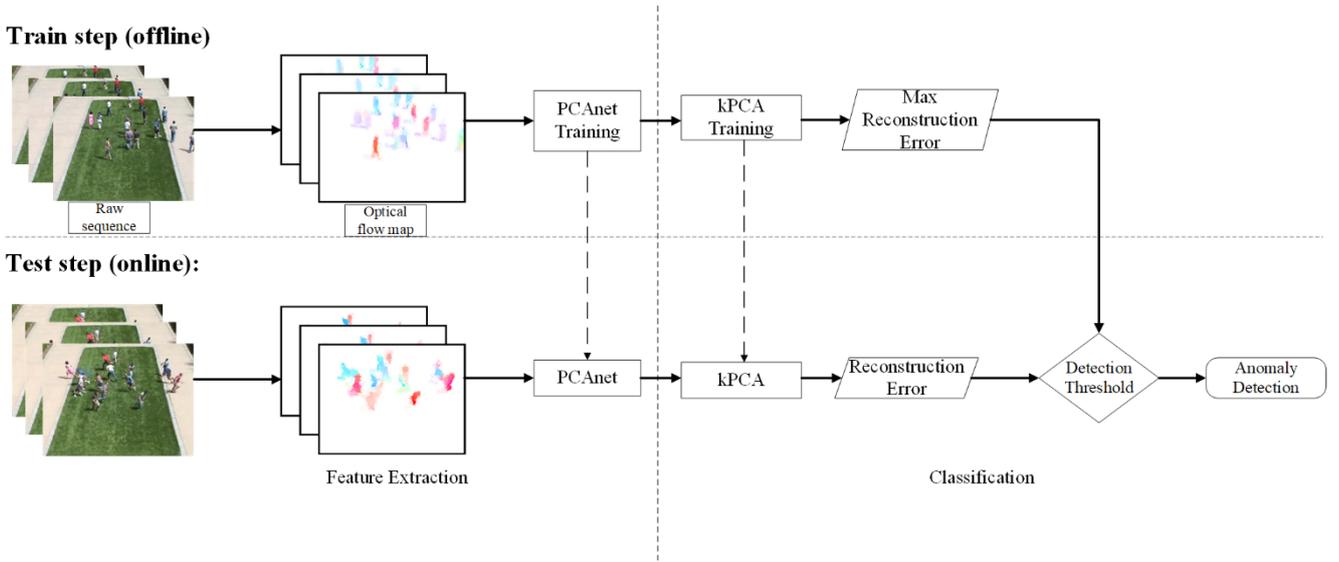

**Fig. 3.** the architecture of the whole framework

4.3. Improved PCAnet with normalization trick

In machine learning field, the generalization of an algorithm is an important but difficult task, which measures the algorithm's performance on new data. Nowadays, the most popular and effective normalization trick in deep learning field is batch normalization(BN)[39]. BN improves network's generalization ability in the way that, giving a sample as input, the output is determined by a whole mini-batch, thus it never produces a deterministic output for a sample. BN's role to elevate model's generalization ability has further been proved by experiments[39]. However, BN is not applicable to our self-supervised model because it has two trainable parameters in the implementation, γ and β. In AED-Net, we could not find ways to train these parameters. Besides, we will not feed data by mini-batches in our method. But, local response normalization (LRN), a light-weight normalization trick with no trainable parameters, is applicable to our task and achieved good results in the experiments.

Proposed by Krizhevsky *et al.*[40], LRN scheme is found to aid generalization ability of the model. Response competition among contiguous outputs having the same spatial position is introduced. For an output value $p_i$ on the $i$-th feature map, the normalized output $q_i$ can be calculated as,

$$q_i = \frac{p_i}{\left(\alpha + \delta \sum_{j \in nb(i,n)} (p_j)\right)^\theta}$$
$$nb(i, n) = \left\{j \Big| j = \max\left(0, \frac{i-n}{2}\right) \ldots \min\left(N-1, \frac{i+n}{2}\right)\right\} \quad (16)$$

where $\delta, n, \alpha, \theta$ are configurable parameters. $\delta$ denotes the weights on outputs of adjacent frames, $\alpha$ is the bias term for computational safety, $\theta$ controls the total magnitude of the normalization term, and $n$ denotes how many adjacent frames are included in the normalization. The feature maps of a network will be arranged once the network is initialized.

We introduce this scheme from CNN to PCAnet to help improve model's ability to generalize. It will be added after computing feature maps by convolution operation on each stage. In addition, LRN's parameters are all set intuitively before training, and they are not learnable, making it suitable for our unsupervised framework.

## 5 Experiments

We carry out experiments on UMN[41] dataset and UCSD Ped1 and Ped2[42] datasets for local abnormal event detection. These public datasets that are open to all of research community are used to evaluate the proposed AED-Net with different criteria: *frame-level criterion* and *pixel-level criterion*, in which UMN dataset evaluate the model's capacity in frame-level, UCSD Ped1 in pixel-level, frame-level and Ped2 in pixel-level. Both evaluation criteria are based on truth-positive rates (TPR) and false-positive rates (FPR), denoting "abnormal events" as "positive" while "normal status" as "negative". The results of experiments are compared with other state-of-art methods, demonstrating superiority over other methods.

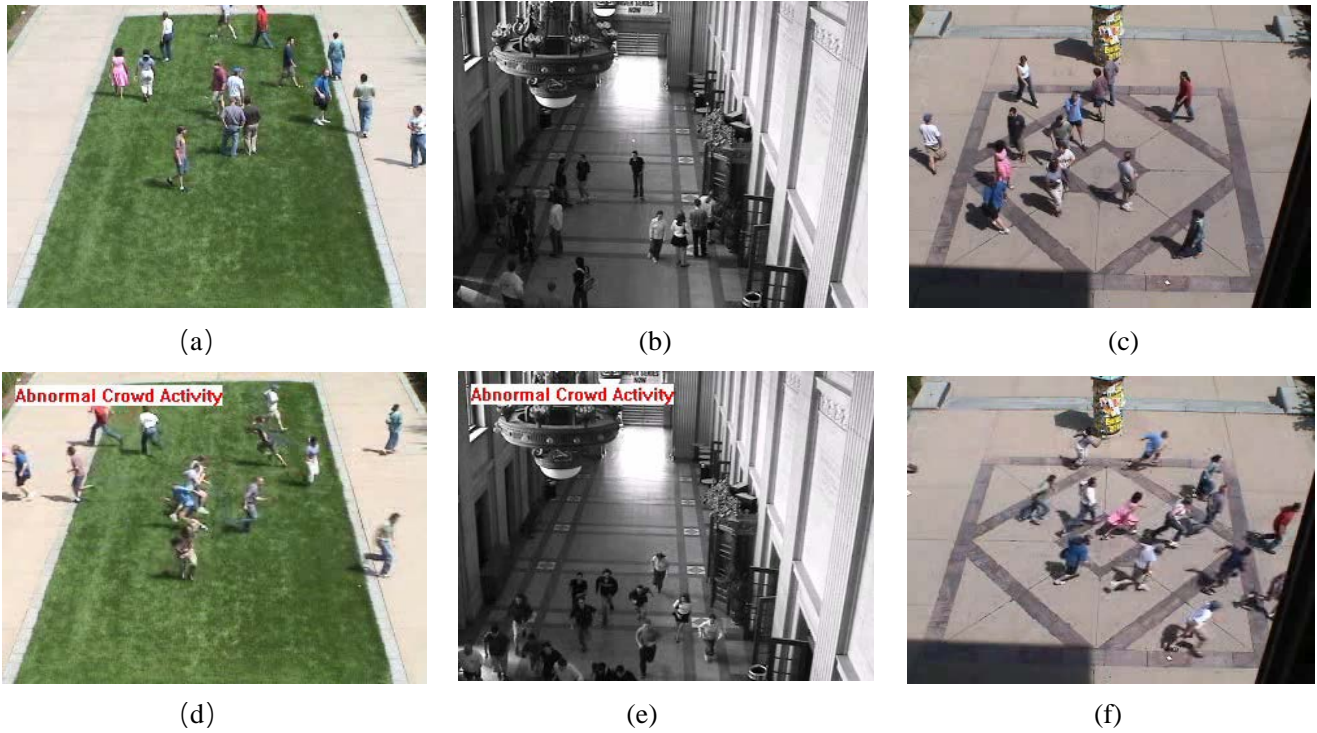

**Fig. 4.** Examples of video frame in three scenes. (a), (d) are examples from scene on lawn, (b), (e) from scene indoor and (c), (f) from scene on plaza

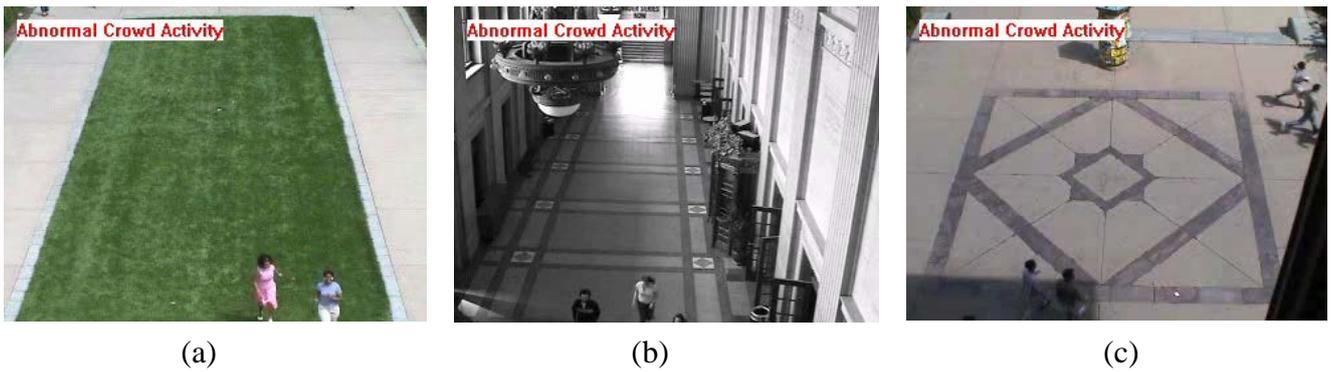

**Fig. 5.** Examples of abnormal video frames detected by considering the area of foreground in the fame due to its disturbance to detection. Similarly, (a) from scene on lawn, (b) from scene indoor and (c) from scene on plaza.

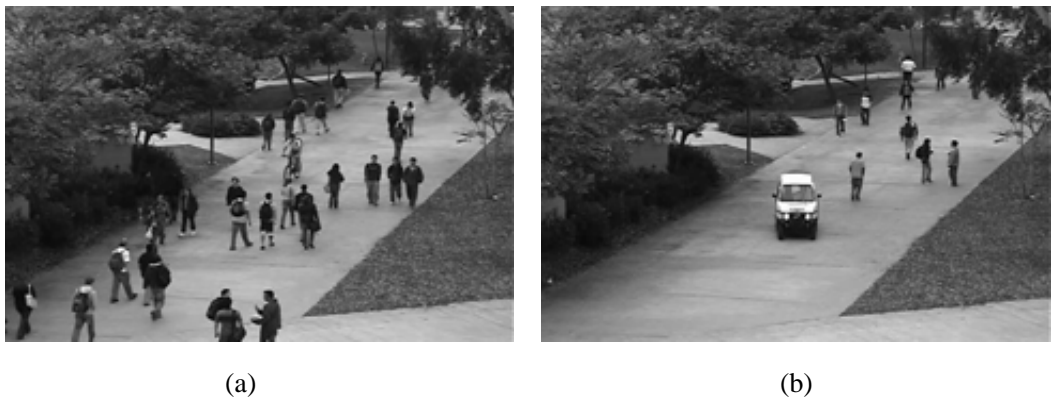

**Fig. 6.** Examples of frames of video clips containing an anomaly. (a) from video clip with anomaly of biker and (b) from video clip with anomaly of a cart

### 5.1 Detection performance on UMN dataset

The UMN dataset [41] is composed of three scenes, lawn, indoor and plaza with a resolution of $240 \times 320$. All scenes are related to escaping action of crowds. In this dataset [41], evacuation behaviors of crowds were assigned as abnormal. We detect the anomalism of each frame so this is measured by frame-level criteria. Fig.4 shows a couple of frames from each of these scenes. For computation efficiency, all optical flow maps extracted from original video frames are resized to small sizes, which had been proved to contain enough information for detection.

Foreground detection is used in this experiment to avoid the disturbance of no-meaning frames. The frames which contain less than three whole human body motion shapes which are shown in Fig.5 are detected directly in our work by measuring moving foreground blobs.

To improve the generalization ability of the AED-Net, data augmentation technique is adopted in this experiment. An Optical flow map is firstly resized to $120 \times 160$ and nine sub-maps with the size of $96 \times 128$ are cut from the resized map. Then, all ten maps (one of $120 \times 160$ and nine of $96 \times 128$) are resized uniformly to $24 \times 32$ for training and testing.

After removing interfering frames, we construct a training set and test set for each scene. 760 normal frames in scene on lawn are used for training, which forms a training set of a size of 7600, other normal and abnormal frames are used for testing. For the scene indoor and on plaza, the number of frames for training are 1100 and 1000.

For all three scenes, the hyper-parameters in AED-Net are set as: filter at each stage had size $3 \times 3$. And both stages had 8 filters that reserve enough variance. The final block size is $8 \times 8$. Hyper-parameters in classifier, kernel size $\sigma$, and number of filters q, differ from each scene. They are set (1,2800), (0.25,4200) and (1,3800) for ($\sigma$, q) for scene on lawn, on plaza and indoor after cross-validation. ROC curve, area under curve (AUC) and equal error rate (EER) are analyzed at frame-level. When plotting ROC curve, the threshold for determining anomalism of frames is altered. The results and comparisons with other methods are presented in Table 2. As table 2 presents, our method achieves respectable results on frame-level anomaly detection both measured by AUC and by EER. Given that the simplicity of whole framework, this result is remarkable, which is better than the state-of-the-art methods.

**Table 2** Results comparison on UMN dataset

| Method | AUC (%) | EER (%) |
| --- | --- | --- |
| Li *et al.*[9] | 99.5 | 3.7 |
| Chaotic invariants[19] | 99.4 | 5.3 |
| Social force[32] | 94.9 | 12.6 |
| sparse[30] | 99.6 | 2.8 |
| Bao *et al.*[43] | - | 2.6 |
| **Ours** | **99.7** | **2.4** |

5.2 Detection performance on UCSD dataset

UCSD dataset[42] contains video clips with resolution of $158 \times 238$, obtained from the camera hung above watching at pedestrian walkways. There are 34 training samples and 36 test samples in Ped1 scene, 16 training samples and 12 testing samples in Ped2 scene of walking people towards different directions were included in it. The video clips labeled as abnormal has one of anomalies including cart, biker, and so on. One of frame with cart anomaly is shown in Fig.6. Then each video frame is partitioned into patches with the size of $12 \times 16$, which contains part of either walking people or anomaly. Then these patches are utilized as raw data. Assigning the anomalism of these patches is called anomaly detection on pixel-level criteria because it is about to classify the abnormality of a different part of pixels of a frame

Similar to previous experiments, foreground detection is also performed here to avoid disturbance. After that, normal patches from video frames containing anomaly of a biker is used as training set, and abnormal patches from two frames of two video clips are used as test set. The hyper-parameters in AED-Net are set as: $k_1 = k_2 = 5$, $L_1 = L_2 = 7$, and block size $7 \times 7$ for experiments. The hyper-parameters in the kPCA classifier are set as: (0.8, 1350) for ($\sigma$, q).

Ped1 pixel-level and frame-level results and comparison with other are shown in Fig.7 and Table 3. Ped2 pixel-level and frame-level results are shown in Table 4. In all experiments, the proposed framework outperforms the state-of-art methods, especially in AUC.

Table 3 Results comparison on UCSD Ped1 scene

| Method | AUC (%) | | EER (%) | |
|---|---|---|---|---|
| | Pixel-level | Frame-level | Pixel-level | Frame-level |
| Li *et al*.[9] | 44.1 | 83.8 | 55.0 | 24.4 |
| MPCCA[42] | 20.5 | 79.6 | 71.8 | 32.9 |
| CDAE[24] | 65.8 | 82.9 | 36.9 | 26.8 |
| sparse[30] | 46.1 | **90.1** | 53.7 | **18.6** |
| SF[32] | 17.9 | 67.0 | 79.0 | 39.2 |
| AED-Net | 86.1 | 88.2 | 22.9 | 22.6 |
| AED-Net + LRN | **88.9** | 89.7 | **19.4** | 19.1 |

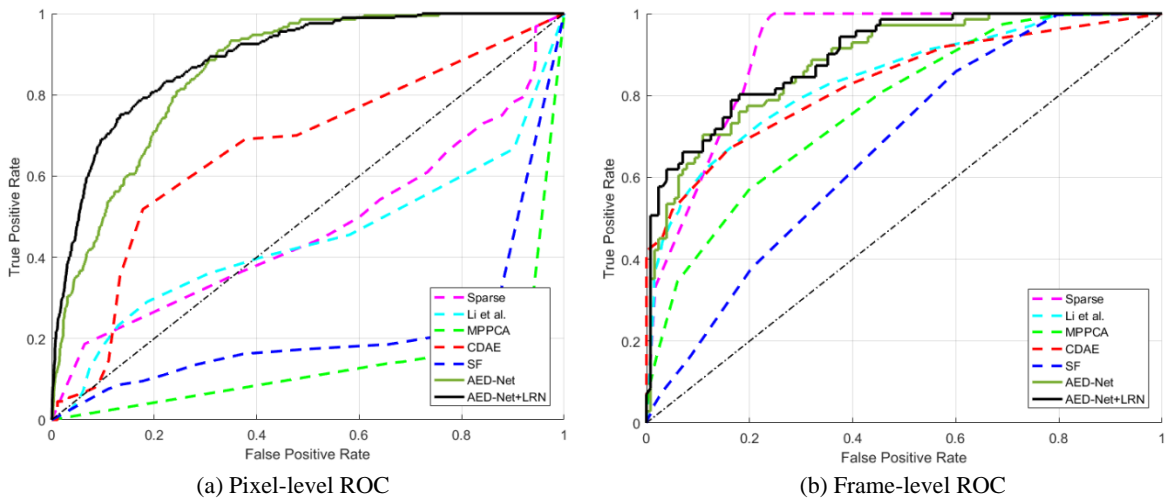

(a) Pixel-level ROC  (b) Frame-level ROC

**Fig. 7.** Results of Ped1 scene. **a.** Pixel-level ROC for Ped1 **b.** Frame-level ROC for Ped1

Table 4 Results comparison on UCSD Ped2 scene

| Method | AUC (%) | | EER (%) | |
|---|---|---|---|---|
| | Pixel-level | Frame-level | Pixel-level | Frame-level |
| Li *et al*.[9] | 70.0 | 85.2 | 29.3 | 18.2 |
| MPCCA[42] | 23.5 | 77.6 | 71.2 | 30.4 |
| SF[42] | 29.1 | 71.6 | 80.2 | 42.3 |
| AED-Net | 88.9 | **90.2** | 22.4 | 20.3 |
| AED-Net + LRN | **91.3** | 89.6 | **16.8** | **15.9** |

5.3 Experiments on improved AED-Net

After adding LRN layer to PCAnet, the whole framework is tested on UCSD dataset, and the experiment setting is the same as the previous one on UCSD dataset. The hyperparameters of LRN is set as, $\gamma = 2, \delta = 10^{-4}$, $n = 5, \beta = 0.75$.

The results shown in Fig.7 and Table 3,4 shows that after adding LRN, the whole framework shows better performance in detecting anomaly both measured by AUC and EER. It indicates that this strategy does improve our method by promoting its generalization ability.

## 6 Conclusion

We propose a simple but efficient framework, AED-Net, based on self-supervised learning method. Raw data from surveillance video clips are used to calculate optical flow maps; then their high-level features are extracted by PCAnet, which are further used to determine the anomalism of the local abnormal events and global abnormal events. From the results in the experiments we can see that the framework performs well on detecting both global abnormal event and local abnormal event. Further, after the LRN layer is added to address overfitting problem, the performance of this framework becomes better. That the framework achieves results better than the state-of-the-art methods indicates that it can effectively extract motion pattern from raw videos and detect anomaly due to it.

**Acknowledgment**
This work is partially supported by the National Key R&D Program of China (2016YFE0204200), the National Natural Science Foundation of China (61503017), the Fundamental Research Funds for the Central Universities (YWF-18-BJ-J-221), the Aeronautical Science Foundation of China (2016ZC51022), and the Platform CAPSEC (capteurs pour la sécurité) funded by Région Champagne-Ardenne and FEDER (fonds européen de développement régional).